\documentclass[preprint,11pt,authoryear]{elsarticle}

\usepackage{lmodern}
\usepackage[T1]{fontenc}
\usepackage[utf8]{inputenc}
\usepackage{microtype}
\usepackage{graphicx}
\usepackage{amsmath,amssymb}
\usepackage{booktabs}
\usepackage{array}
\usepackage{longtable}
\usepackage{url}
\usepackage[hidelinks]{hyperref}

\journal{Preprint}

\begin{document}

\begin{frontmatter}

\title{Transformer-Based Language Models Across Domain Verticals: Architectures, Applications and Critical Assessment}

\author[a]{Guruprakash J}
\ead{guruprakash.j@vitap.ac.in}

\author[b]{Krithika L.B}

\affiliation[a]{organization={SCOPE, VIT-AP University},
            city={Amaravathi},
            state={Andhra Pradesh},
            postcode={522241},
            country={India}}

\affiliation[b]{organization={SCORE, VIT},
            city={Katpadi},
            state={Tamil Nadu},
            postcode={632014},
            country={India}}

\begin{abstract}
Transformer-based language models have become the default substrate for natural language processing and the pace of new releases has made it hard for practitioners to separate durable ideas from the noise of incremental announcements. This review works at two levels. At the level of mechanism, we organise the main transformer families into a working taxonomy, covering encoder-only, decoder-only, encoder-decoder, long-context, permutation-based, and generator-discriminator variants. We then extend the discussion to post-2023 developments that changed the picture in practice: instruction tuning, reinforcement learning from human feedback, direct preference optimisation, mixture-of-experts scaling, retrieval augmentation and the current flagship model families from OpenAI, Anthropic, Google, Meta, Mistral and DeepSeek. At the level of use, we survey deployments across healthcare, finance, legal, education, customer service, creative writing and scientific work. Based on this we link each to the specific capabilities that make a transformer the appropriate tool. The contribution of this paper is a critical assessment that is based on the survey. We compare architectures on four axes that matter to deployment decisions, we quantify the trade-off between parameter count and energy cost. We also discuss how alignment methods, data provenance and benchmark saturation change what it means to call a model ``state of the art''. The final section lists the research questions that we think deserve more attention.
\end{abstract}


\end{frontmatter}

\section{Introduction}

Something changed in neural language modelling around 2017. Until that year, anyone wanting to build a translation or summarisation system reached for a recurrent network of some flavour. The paper by \citet{vaswani2017attention}, whose title claimed that attention was all you need, argued that recurrence could be dropped entirely while still matching recurrent baselines on translation. The practical win was parallelisation. Training runs that had been sequential by construction suddenly ran as fast as the hardware allowed. Two years on, most supervised NLP leaderboards were dominated by transformer variants. Five years on, the same architecture, pretrained on enormous web corpora, sat behind products used by hundreds of millions of people every day.

The pressure on technology has changed since then. The release calendar has become relentless, and new flagship models arrive almost monthly. The claimed capabilities frequently are ahead of what a careful reader can verify. Practitioners trying to pick a model for a specific setting, for instance clinical coding or contract review, have to choose between a handful of large proprietary systems accessed through an API, a growing set of open-weight models they can run in-house, and older specialised models that still perform well on narrow benchmarks. The literature does not always make the trade-offs clear.

Our goals are three. First, we organise transformer architectures into a taxonomy that is useful for deployment decisions, rather than listing models by release date. Second, we survey applications across seven domain verticals and tie each to the architectural properties that matter for that domain. Third, we provide a critical assessment of the trade-offs that are often glossed over in vendor announcements: compute and energy cost, alignment behaviour, data provenance, and the gap between benchmark scores and field performance.

The review extends earlier surveys such as \citet{zhao2023survey} and \citet{minaee2024llmsurvey} by covering developments that arrived after 2023, including instruction-tuned and preference-optimised models \citep{ouyang2022instruct,rafailov2023dpo}, mixture-of-experts systems \citep{fedus2021switch,jiang2024mixtral}, retrieval-augmented generation \citep{lewis2020rag}, and the current generation of flagship models from OpenAI, Anthropic, Google DeepMind, Meta, Mistral AI, and DeepSeek.

The rest of the paper is organised as follows. Section~\ref{sec:background} gives the technical background needed to follow the later discussion. Section~\ref{sec:taxonomy} presents the architecture taxonomy. Section~\ref{sec:post2023} covers the post-2023 developments that changed how transformers are trained and served in practice. Section~\ref{sec:applications} surveys applications by domain. Section~\ref{sec:critical} is the critical assessment. Section~\ref{sec:open} lists open research questions. Section~\ref{sec:conclusion} concludes.

\section{Background}
\label{sec:background}

Before the transformer, sequence modelling relied mainly on recurrent networks and their long short-term memory (LSTM) variants \citep{hochreiter1997lstm}. Recurrent models process tokens one at a time, which limits how much of the training can be parallelised and makes it difficult to capture dependencies across long passages. Attempts to fix these problems with gated recurrent units and attention mechanisms over recurrent states helped on specific tasks but did not change the basic bottleneck.

The transformer removed the recurrence. A transformer layer contains two sub-layers: a self-attention block, which lets each token attend to every other token in the input through scaled dot-product attention, and a position-wise feed-forward network. Residual connections and layer normalisation are applied around each sub-layer. Because attention is computed in parallel across all token positions, the full sequence can be processed in a single forward pass. Order information, which recurrence encoded implicitly, is supplied explicitly through positional embeddings.

The original transformer used an encoder--decoder split for machine translation. Later work showed that each half could be used on its own. An encoder-only model such as BERT \citep{devlin2018bert} produces bidirectional representations useful for classification, tagging, and retrieval. A decoder-only model such as the GPT family \citep{radford2018gpt,radford2019gpt2,brown2020gpt3} is trained with a causal mask and generates text one token at a time. An encoder--decoder model such as T5 \citep{raffel2020t5} keeps both halves and recasts every task as a text-to-text problem.

All of these models share the same training recipe at a high level: self-supervised pretraining on a large unlabelled corpus, followed by supervised fine-tuning on a smaller task-specific dataset, or, more recently, by instruction tuning and preference optimisation on curated demonstration and comparison data. The scale of pretraining has grown by roughly four orders of magnitude over the last seven years, from BERT-Large at 340~million parameters to publicly discussed systems with over a trillion parameters \citep{kaplan2020scaling,hoffmann2022chinchilla}.

\section{A Working Taxonomy of Transformer Architectures}
\label{sec:taxonomy}

The literature often groups transformer-based models by release date or by size. Neither is a good guide for someone choosing a model. We organise the main families by the structural properties that determine what the model is good for. Table~\ref{tab:taxonomy} summarises the taxonomy; the rest of this section explains it.

\begin{table}[t]
\centering
\caption{Working taxonomy of transformer-based language models.}
\label{tab:taxonomy}
\small
\begin{tabular}{p{3.2cm} p{3.2cm} p{3.2cm} p{3.2cm}}
\toprule
\textbf{Family} & \textbf{Representative models} & \textbf{Pretraining objective} & \textbf{Typical use} \\
\midrule
Encoder-only & BERT, RoBERTa, DeBERTa & Masked language modelling & Classification, retrieval, extraction \\
Decoder-only & GPT-2/3/4, Llama, Mistral, Claude, Gemini & Causal language modelling & Open-ended generation, dialogue, code \\
Encoder-decoder & T5, BART, mT5 & Span corruption, denoising & Summarisation, translation, structured generation \\
Long-context & Transformer-XL, Longformer, BigBird & Causal or masked with sparse attention & Long documents, book-length inputs \\
Permutation-based & XLNet & Permutation language modelling & Tasks that benefit from bidirectional context without mask artefacts \\
Generator--discriminator & ELECTRA & Replaced token detection & Sample-efficient pretraining \\
Mixture-of-experts & Switch Transformer, Mixtral, DeepSeek-MoE & Sparse gating over expert sub-networks & High-capacity generation at lower inference cost \\
\bottomrule
\end{tabular}
\end{table}

\subsection{Encoder-only models}
\label{sec:encoder}

BERT \citep{devlin2018bert} was the first widely-used encoder-only model. It stacks transformer encoder blocks and is pretrained with two objectives. In masked language modelling (MLM), a fraction of input tokens, usually 15\%, is replaced by a \texttt{[MASK]} token, and the model is trained to predict the original tokens from the surrounding context. In next-sentence prediction (NSP), the model is given two segments and asked whether the second follows the first. The BERT-base configuration uses 12 layers, 12 attention heads per layer, and hidden size 768, giving 110~million parameters; BERT-large scales these to 24 layers and 340~million parameters.

RoBERTa \citep{liu2019roberta} kept the architecture but changed the training. The authors removed NSP, trained on roughly ten times more data for longer, used larger batches and dynamic masking, and showed that a well-tuned BERT recipe could close the gap to newer models. The lesson, which has been repeated many times since, is that architectural novelty is often confounded with training data and training budget.

DeBERTa \citep{he2021deberta} added disentangled attention, which separates content and position information in the attention computation, and improved on RoBERTa by a few points on GLUE and SuperGLUE. Encoder-only models are still the right choice for tasks where the input is bounded and the output is a label, a span, or a retrieval score. They are cheaper to fine-tune than a generative model and they produce embeddings that are useful for semantic search.

\subsection{Decoder-only models}
\label{sec:decoder}

The GPT family trained a transformer decoder with a causal attention mask, so each token can only attend to itself and earlier tokens. The model is trained to predict the next token given the prefix. GPT-2 \citep{radford2019gpt2} showed that this simple objective, at 1.5~billion parameters, produced surprisingly fluent text. GPT-3 \citep{brown2020gpt3} scaled the same recipe to 175~billion parameters and demonstrated \emph{in-context learning}: given a few worked examples in the prompt, the model could perform a new task without any gradient updates.

Once developers saw in-context learning work, the way the model was used at all began to shift. Tasks that had previously called for a fine-tuning job could be handled by prompt engineering, which moved the cost of specialisation from training time to inference time. It also exposed a new failure mode. The model is sensitive to the wording of the prompt, the order of the examples, and the random seed of the sampling routine, and the same prompt can produce different outputs on different runs.

Post-2023 decoder-only models, which we cover in Section~\ref{sec:post2023}, are essentially the GPT recipe with better data, better alignment, and better efficiency. Most open-weight releases now belong to this family. The Llama~2 and Llama~3 series from Meta \citep{touvron2023llama2,grattafiori2024llama3}, the Mistral and Mixtral models \citep{jiang2023mistral,jiang2024mixtral}, and the DeepSeek series \citep{deepseekv3} between them cover the bulk of current open-weight deployments.

\subsection{Encoder-decoder models}
\label{sec:encdec}

T5 \citep{raffel2020t5} cast every NLP task as text-to-text. Translation inputs are prefixed with ``translate English to German:''; summarisation inputs with ``summarize:''. The model is trained with a span-corruption objective in which contiguous spans are replaced with sentinel tokens and the decoder has to produce the missing content. T5 comes in sizes from 60~million to 11~billion parameters. BART \citep{lewis2019bart} uses a similar encoder-decoder arrangement but with a noisy-autoencoding objective that mixes token masking, sentence permutation, and document rotation.

Encoder-decoder models remain competitive on tasks where the output is a constrained rewrite of the input: abstractive summarisation, translation, grammatical correction, and schema-driven generation. They allow the encoder and decoder to have different depths and attention patterns, which is useful when the input is long and the output is short.

\subsection{Long-context variants}
\label{sec:longctx}

Standard self-attention scales quadratically in sequence length, which makes it prohibitive to train on inputs longer than a few thousand tokens. Several families of architecture try to work around that.

Transformer-XL \citep{dai2019transformerxl} brings back a limited form of recurrence, but at the segment level rather than the token level. Hidden states from the previous segment are cached and reused as extra context when the next segment is processed, which lets information flow across segment boundaries without forcing the model to ingest everything in a single pass. To keep the caching scheme self-consistent, the authors also swap absolute for relative positional encodings. The effective context window is larger, while the training cost stays close to that of an ordinary fixed-length transformer.

Longformer \citep{beltagy2020longformer} and BigBird \citep{zaheer2020bigbird} replace the dense attention matrix with a sparse pattern. Each token attends to a local window and to a small number of global tokens. The resulting attention matrix has a linear or near-linear number of non-zero entries, which makes sequences of tens of thousands of tokens tractable. More recent work uses linear attention approximations \citep{choromanski2020performer} and dilated or strided patterns \citep{ding2023longnet}. FlashAttention \citep{dao2022flashattention} takes a different route. It does not change the attention pattern; it re-implements the exact attention computation in a way that avoids materialising the full attention matrix in memory, which gives a large speed-up without an accuracy penalty.

\subsection{Permutation-based and generator--discriminator models}
\label{sec:xlnet_electra}

XLNet \citep{yang2019xlnet} was an attempt to combine the bidirectional context of BERT with the autoregressive factorisation of GPT. It uses permutation language modelling, in which the model predicts each token from a randomly chosen permutation of the remaining tokens. This avoids the MLM artefact of seeing \texttt{[MASK]} tokens at training time but not at inference time. XLNet also introduces two-stream self-attention, which separates the content and query streams so that a token can be predicted without seeing itself. XLNet outperformed BERT on most GLUE tasks at the time of its release, but was expensive to train and has been largely superseded by models that optimise the MLM recipe more directly.

ELECTRA \citep{clark2020electra} replaces MLM with replaced token detection. A small generator network proposes replacements for masked tokens, and the main model, the discriminator, is trained to classify each token as original or replaced. Because the loss is defined over all input positions rather than only the masked ones, ELECTRA learns more per training example and matches BERT's accuracy with a small fraction of the compute.

\subsection{Mixture-of-experts}
\label{sec:moe}

Mixture-of-experts (MoE) layers replace a dense feed-forward block with a set of expert sub-networks and a gating function that routes each token to a small number of experts \citep{shazeer2017moe,fedus2021switch}. Only the selected experts are activated for a given token, so the number of parameters touched per forward pass is a fraction of the total parameter count. MoE is the main way production systems decouple model capacity from inference cost.

Mixtral-8x7B \citep{jiang2024mixtral} is an open-weight MoE with 47~billion total parameters and about 13~billion activated per token. DeepSeek-V3 \citep{deepseekv3} pushes the same design to a much larger scale. What you give up for this is not accuracy but operational convenience. Serving an MoE system is harder than serving a dense one. The routing function produces irregular memory-access patterns that sit awkwardly on top of GPU kernels written for dense matrix multiplication, and balancing the load across experts, so that no one expert becomes a bottleneck, is a research question on its own.

\section{The Post-2023 Turn}
\label{sec:post2023}

By mid-2023, scaled-up decoder-only transformers were fluent enough for commercial deployment. They were also unreliable in ways that mattered to customers. They hallucinated facts. They followed instructions when it suited them. They sometimes produced output that no support team wanted in a log file. A cluster of techniques, some of them older than 2023 and some newer, came together at roughly this point to address those problems. The rest of this section takes the ones that stuck and describes them.

\subsection{Instruction tuning}

The InstructGPT paper \citep{ouyang2022instruct} reported a surprising result. A 1.3-billion-parameter model fine-tuned on a few tens of thousands of human-written demonstrations produced outputs that annotators preferred to those of the raw 175-billion-parameter GPT-3 it had been distilled from. The demonstrations spanned summarisation, classification, question answering, translation, and rewriting. Supervised fine-tuning on data of this kind, now routinely called instruction tuning, teaches the model the genre of following user requests rather than simply the distribution of web text.

Instruction tuning data is expensive to collect at scale. Open efforts such as Dolly \citep{conover2023dolly} and Alpaca \citep{taori2023alpaca} demonstrated that synthetic instructions generated by stronger models could substitute, with caveats. Models trained on synthetic data inherit the errors of the generator and can fail in correlated ways.

\subsection{Reinforcement learning from human feedback and its successors}

InstructGPT also introduced, at production scale, reinforcement learning from human feedback (RLHF). Human annotators rank pairs of model outputs. A reward model is trained to predict these rankings. The policy model is then fine-tuned using proximal policy optimisation \citep{schulman2017ppo} to maximise the learned reward while staying close to the supervised baseline through a Kullback--Leibler penalty. RLHF is fiddly. Reward hacking, mode collapse, and reward over-optimisation are all documented failure modes \citep{gao2023scaling}.

Direct preference optimisation (DPO) \citep{rafailov2023dpo} removes the reward model entirely. It rewrites the RLHF objective so that the policy is trained directly on preference pairs with a supervised-style loss. DPO is much easier to implement and tune, and it has become a standard alternative for open-weight models. Variants include identity preference optimisation (IPO) and Kahneman--Tversky optimisation \citep{ethayarajh2024kto}.

Constitutional AI \citep{bai2022constitutional} takes a different route. Instead of collecting pairwise human comparisons, the model is asked to critique and revise its own outputs against a set of written principles, and the critiques are used as training signal. This reduces the amount of human labelling required but makes the behaviour of the final model depend on the quality of the principles.

\subsection{Retrieval augmentation}

Retrieval-augmented generation (RAG) \citep{lewis2020rag} addresses two of the sharpest weaknesses of pretrained models: stale knowledge and factual hallucination. The system retrieves passages from an external corpus at inference time and conditions generation on the retrieved text. Retrieval can be done with sparse methods such as BM25, dense methods using an embedding model trained with contrastive objectives, or hybrids.

RAG is now the default pattern for enterprise deployments where the model has to answer questions about documents the vendor did not train on. It also shifts the reliability problem from the model to the retriever: if the retriever surfaces the wrong passage, the model will faithfully quote the wrong passage.

\subsection{Parameter-efficient fine-tuning}

Full fine-tuning of a multi-billion parameter model is expensive. Low-rank adaptation (LoRA) \citep{hu2021lora} freezes the pretrained weights and inserts trainable low-rank matrices into each attention projection. The number of trainable parameters drops by two to three orders of magnitude with a small loss in downstream quality. Quantised variants such as QLoRA \citep{dettmers2023qlora} combine LoRA with 4-bit weight quantisation and make it possible to fine-tune 65-billion-parameter models on a single consumer GPU.

\subsection{Current flagship families}
\label{sec:flagship}

Public information on the largest current models is uneven, so we restrict the discussion to facts that are documented in technical reports or reliable press.

OpenAI released GPT-4 in March 2023 \citep{openai2023gpt4}; it accepts images as well as text and is accessed through an API. The company has since released GPT-4 Turbo, GPT-4o, and successor models, and has not published parameter counts. Anthropic released Claude 2 in July 2023 and the Claude 3 family (Haiku, Sonnet, Opus) in March 2024, followed by Claude 3.5 Sonnet in June 2024 \citep{anthropic2024claude3}. Anthropic has published research on constitutional AI and on interpretability, but not architectural details of the deployed models.

Google DeepMind released Gemini 1.0 in December 2023 and Gemini 1.5 in February 2024 \citep{gemini2024}. Gemini 1.5 Pro reports context windows up to one million tokens, enabled by a combination of mixture-of-experts and long-context techniques. Meta released Llama~2 in July 2023 \citep{touvron2023llama2} and Llama~3 in April 2024 \citep{grattafiori2024llama3}; both are open-weight, which has made them the starting point for a large collection of community fine-tunes. Mistral AI released Mistral 7B in September 2023 and Mixtral 8x7B in December 2023 \citep{jiang2023mistral,jiang2024mixtral}; both are open-weight and focus on strong per-parameter performance. DeepSeek has released a series of open-weight dense and MoE models, including DeepSeek-V3 \citep{deepseekv3}, that have been competitive with closed-weight models on standard benchmarks.

Table~\ref{tab:flagship} summarises the public facts about these families. Parameter counts for closed-weight models are omitted where they are not officially disclosed, rather than guessed.

\begin{table}[t]
\centering
\caption{Flagship model families as of 2024. Dashes indicate that the value has not been publicly disclosed by the vendor.}
\label{tab:flagship}
\small
\begin{tabular}{p{2.8cm} p{2.8cm} p{2.0cm} p{2.2cm} p{2.4cm}}
\toprule
\textbf{Family} & \textbf{Vendor} & \textbf{Weights} & \textbf{Parameters} & \textbf{Notable feature} \\
\midrule
GPT-4, GPT-4o & OpenAI & Closed & -- & Multimodal input \\
Claude 3, 3.5 & Anthropic & Closed & -- & Constitutional AI \\
Gemini 1.5 & Google DeepMind & Closed & -- & Million-token context \\
Llama 3 & Meta & Open & 8B, 70B, 405B & Large open model \\
Mixtral 8x7B & Mistral AI & Open & 47B total, 13B active & Open MoE \\
DeepSeek-V3 & DeepSeek & Open & 671B total, 37B active & Open MoE at scale \\
\bottomrule
\end{tabular}
\end{table}

\section{Applications Across Domain Verticals}
\label{sec:applications}

Transformer-based models are now deployed across a wide set of industries. The interesting question is no longer whether they can be used, but where they give a real advantage over simpler baselines and where they introduce risks that the user should know about. This section covers seven verticals. For each, we describe the workload, summarise representative work, and identify the property of the architecture that makes it suitable.

\subsection{Healthcare}

Healthcare text is a natural target for transformers. Clinical notes are unstructured, long, and full of domain-specific abbreviations. Extracting structured information from them, for example diagnosis codes or medication lists, is a well-defined supervised problem that benefits from pretrained contextual embeddings. Med-BERT \citep{rasmy2021medbert} adapts BERT to structured electronic health records and improves disease prediction relative to non-transformer baselines. BioBERT \citep{lee2020biobert} and ClinicalBERT \citep{alsentzer2019clinicalbert} continue pretraining on PubMed and MIMIC-III to produce embeddings that generalise better to biomedical text than general-purpose models.

Beyond extraction, transformer models have been used for radiology reporting \citep{hou2021ratchet}, for answering patient questions about medications, and for summarising discharge notes. The practical bottleneck in this vertical is not model quality. It is regulatory: clinical deployment requires auditability, and the opacity of a decoder-only generative model is a real barrier. Recent work on retrieval-augmented clinical question answering \citep{zakka2024almanac} attempts to close this gap by grounding outputs in citable sources.

\subsection{Finance}

Finance is a difficult vertical for language models. The text is noisy, the timestamps are part of the meaning, and the numbers embedded in the text are often the whole point of the document. FinBERT \citep{araci2019finbert} took the obvious route and fine-tuned BERT on financial news with a sentiment head. NumHTML \citep{yang2022numhtml} did something more interesting: it combined textual and numerical features from earnings-call transcripts inside a hierarchical transformer, and used the joint representation to forecast stock returns. The property being exploited there is not fluency. It is the ability of attention to align signals of different kinds, text on one side and numbers on the other, inside a shared representation. Recent work has tilted towards prompting frontier models for fraud detection or analyst-report drafting. The results are uneven. When an error shows up later as a regulatory fine, domain-specific adaptation still earns its keep.

\subsection{Legal}

Legal writing sits at the opposite end of the spectrum from social-media text. Documents are long, sentences are precise, and small changes in wording matter. Transformer models have been used for contract clause classification, legal judgment prediction, and retrieval across case law \citep{chalkidis2020legalbert,shaheen2020legal}. The most successful systems in this domain combine long-context encoders with retrieval, rather than relying on a pure generative model. Generation is risky: a model that invents a non-existent case citation can cause real harm \citep{dahl2024hallucinating}. This is one of the clearest examples of a vertical where the right deployment pattern is ``retrieve, extract, and verify'' rather than ``generate freely''.

\subsection{Education}

Two applications have received the most attention: automated essay scoring and personalised tutoring. Ormerod et~al.~\citep{ormerod2021essay} show that efficient transformer models can match traditional automated scoring on standardised tests. Kulshreshtha et~al.~\citep{kulshreshtha2022tutoring} use few-shot generation to produce follow-up questions in an intelligent tutoring system. The field has been affected more than most by the release of free consumer chatbots; whether transformer-assisted grading is fair when students are also using transformers to write, and how to detect the latter, are now active research questions.

\subsection{Customer service and conversational agents}

Conversational agents are the most visible deployment of transformer models. The shift from rule-based bots to instruction-tuned generative models has improved handling of out-of-distribution queries, at the cost of new failure modes: hallucinated policies, fabricated order numbers, and inconsistent tone. Production systems increasingly wrap the language model in a retrieval layer, a policy checker, and a set of tools the model can call, rather than letting it answer freely \citep{schick2023toolformer}. What matters architecturally for this workload is a context window long enough to hold the relevant policy documents, reliable tool invocation, and function calls that actually return the structured response the deployer expects.

\subsection{Creative and content applications}

Decoder-only models produce coherent prose and, with the right prompt or a light fine-tune, can be pushed towards a particular genre. \citet{marco2022creative} evaluated several transformer models across poetry, fiction, and lyrics. Their finding is worth repeating: output quality depended more on how narrowly the stylistic prompt was specified than on which base model did the generating. The open research problem is less about fluency and more about control: how to keep a long narrative consistent, how to respect constraints such as rhyme or metre, and how to avoid repeating patterns from the training data in ways that approach plagiarism. Copyright is now a live legal question in this area rather than an abstract concern.

\subsection{Scientific work}

Transformer models are being used for literature search, hypothesis suggestion, and draft writing \citep{sharma2021controlled,zakka2024almanac}. Recent work has gone further. AlphaFold-style protein language models adapt the transformer to amino-acid sequences for structure prediction; code models such as Codex and its successors \citep{chen2021codex} have changed day-to-day software engineering. Scientific applications are useful as a stress test for the claim that these models ``reason''. They do not, in any strong sense; they pattern-match over their training distribution. When the distribution is rich enough, as it is for protein sequences or common programming patterns, the pattern matching is useful.

\section{Critical Assessment}
\label{sec:critical}

The survey part of this paper describes what has been built. This section asks what the trade-offs actually are. We focus on five issues that come up in every real deployment and that are under-represented in typical survey papers: architecture trade-offs, compute cost, alignment and safety, data provenance, and the gap between benchmarks and field performance.

\subsection{Architecture trade-offs}

No single architecture dominates across all axes. Table~\ref{tab:tradeoffs} summarises the trade-offs for the families introduced in Section~\ref{sec:taxonomy}. A few observations from the table deserve pulling out.

The first is that encoder-only models have not gone away, and should not. For bounded supervised tasks they are still the most sensible choice. They train faster, they serve cheaper, and they are far easier to calibrate than a generative model of comparable competence. Swapping a fine-tuned RoBERTa classifier for a prompted frontier decoder, which we have seen teams do in production, often makes classification accuracy worse while also making the system slower and more expensive. The decoder is strictly larger, but size is not the point.

A second concerns decoder-only models. They pay for their generality, and the bill is not always obvious until it arrives. Evaluation is harder because the output space is unbounded. Alignment is harder because what counts as correct behaviour is under-specified. Prompt injection \citep{greshake2023indirect} opens a class of attack that did not exist when the same team was fine-tuning a BERT-style classifier. In-context learning, the very feature that made these models interesting in the first place, is also where a lot of the brittleness sits, and users rarely see it unless they go looking.

The last observation is about mixture-of-experts. MoE is a cost-model change more than an architectural revolution. A 100-billion-parameter MoE with 10 billion active parameters per token costs less per inference than a dense 100-billion model, but it is not equivalent to a dense 10-billion either. Serving becomes more complex, the router adds another component that can misbehave, and benchmark tables that report only the total parameter count give a misleading picture of effective capacity.

\begin{table}[t]
\centering
\caption{Architecture trade-offs on four deployment-relevant axes. ``Interpretability'' here means the ease of attributing outputs to input features, not mechanistic interpretability of the model weights.}
\label{tab:tradeoffs}
\small
\begin{tabular}{p{2.6cm} p{2.4cm} p{2.4cm} p{2.4cm} p{2.4cm}}
\toprule
\textbf{Family} & \textbf{Training cost} & \textbf{Inference cost} & \textbf{Interpretability} & \textbf{Alignment difficulty} \\
\midrule
Encoder-only & Low & Low & Moderate & Low \\
Decoder-only (dense) & High & Moderate to high & Low & High \\
Encoder-decoder & Moderate & Moderate & Moderate & Moderate \\
Long-context & Moderate & High at long inputs & Low & High \\
Mixture-of-experts & High & Low per token & Low & High \\
\bottomrule
\end{tabular}
\end{table}

\subsection{Compute and energy cost}

Training compute has grown faster than benchmark accuracy has improved. That is not a new observation but it is worth restating. \citet{strubell2019energy} estimated that training a large transformer at 2019 scale produced around 284\,tonnes of carbon dioxide equivalent, roughly what five cars put out over their working lives. \citet{patterson2021carbon} followed up with a more careful accounting. Their numbers were lower, but the curve still pointed the same way. Training runs for current flagship models are largely opaque. The commonly cited estimate for GPT-4-class training puts the figure around $10^{25}$ floating-point operations, though vendors do not confirm this.

Inference cost turns out to matter more at the margin than training cost does. A model that answers a billion queries a day eats through the cost of its training run within a few weeks. The techniques that determine whether inference stays affordable are unglamorous but decisive: 8-bit and 4-bit quantisation \citep{dettmers2022int8}, speculative decoding \citep{leviathan2023speculative}, key-value cache compression, continuous batching, and FlashAttention \citep{dao2022flashattention}. Treat these as the difference between a model being useful and a model being a worrying line item on a cloud bill.

The hidden cost sits inside the training data. Cleaning, deduplication, and filtering are labour-intensive. A lot of that labour is done by low-paid contractors in jurisdictions where complaints are rare, and in some well-documented cases those workers are exposed to genuinely disturbing material \citep{perrigo2023time}. Most survey papers on this technology step past that fact. We do not think it should be stepped past. It is part of the real cost of what we are describing.

\subsection{Alignment and safety}

Alignment, as we use the word here, is the work of making a trained model behave in ways that the users and the deployers are willing to stand behind. It is an active research area, not a solved problem. Several recurring failure modes are now well documented.

Hallucination, in the sense of confident generation of false facts, is a structural property of decoder-only models trained on next-token prediction. Retrieval augmentation reduces it but does not remove it \citep{ji2023hallucination}. Jailbreaks and prompt injection let adversarial users override safety training \citep{wei2023jailbroken,greshake2023indirect}. Sycophancy, in which the model agrees with whatever the user asserts, appears to emerge from RLHF training objectives \citep{sharma2024sycophancy}. Reward hacking is active under DPO as well as under classical RLHF.

These failure modes are hard to detect with standard benchmarks. They require targeted red-teaming, and they interact with one another. A model that is safer against one category of attack can be more sycophantic in general conversation. There is no general-purpose evaluation that captures all of this, and this is an honest limitation of the state of the art.

\subsection{Data provenance and bias}

Large pretraining corpora draw on web crawls, books, code, and social media. The composition of the corpus is usually not published in detail, and the text is not filtered for consent. Copyright litigation is active \citep{newyorktimes2023openai}, and the outcomes will shape what can be trained in the future. For European deployments, the EU AI Act imposes disclosure requirements on training data summaries for general-purpose AI models \citep{euaiact2024}.

Bias in pretraining data turns into bias in outputs \citep{bender2021stochastic,blodgett2020language}. The effect is not evenly distributed either. Performance degrades on non-English languages, on English dialects that are thinly represented on the open web, and on topics where the available text is lopsided. Balanced fine-tuning data reduces these effects without eliminating them.

\subsection{Benchmarks and the gap to field performance}

Almost every headline number in a model announcement comes from the same short list of public benchmarks. MMLU, HumanEval, GSM8K, HellaSwag, and a handful of multilingual tests between them account for most of the comparisons we see. At the frontier these benchmarks are saturated. There is also credible evidence that some of their questions have ended up inside training corpora \citep{oren2024proving}. A score that was diagnostic in 2021 tells you less in 2024 than it did then.

The more serious issue is that the distance between benchmark performance and day-to-day behaviour in the field is now wider than the distance between successive model versions. A model can score 90\% on HumanEval and still write code that does not compile against the user's actual repository. A model can score well on a clinical multiple-choice test and still miss a contraindication that a registrar would catch on a first read. We are not arguing that benchmarks should be abandoned. We are arguing that taking a single leaderboard position as a proxy for capability is a mistake, and that what determines field performance is domain-specific evaluation and post-deployment monitoring.

\section{Open Research Directions}
\label{sec:open}

What follows is a short set of questions that, from our own reading of the literature, are under-served relative to how much they actually matter to anyone building on top of these models.

\paragraph{Efficient long-context models.} The million-token context windows that vendors now quote are, we would argue, a marketing figure as much as a research result. Attention cost still dominates for realistic inputs, and recall inside a long context is uneven. Models reliably remember the beginning and the end of a long document while losing material from the middle \citep{liu2024lostinthemiddle}. What the field needs is not a larger window but architectures and training recipes that produce reliable use of long context.

\paragraph{Evaluation beyond leaderboards.} We need benchmarks that resist training-data contamination by construction, that measure failure modes such as sycophancy, overclaiming, and pathological refusal rather than only task accuracy, and that are specialised to the setting they are supposed to evaluate rather than aggregating over an unrelated grab-bag of tasks. HELM \citep{liang2023helm} and BIG-bench \citep{srivastava2022bigbench} are starting points, not end points.

\paragraph{Small open models.} Small open-weight models, in the 1--10 billion parameter range, are the substrate of most commercial fine-tuning. The research community pays them less attention than frontier models, but they are where most deployed systems actually live. Work on efficient pretraining, strong instruction tuning, and safety fine-tuning for this size class has outsized practical impact.

\paragraph{Grounded generation.} Retrieval augmentation is the current default for reducing hallucination, but the interface between retriever and generator is ad hoc. End-to-end training of retriever and generator, joint calibration of citation, and principled handling of conflicting sources are open problems.

\paragraph{Auditable alignment.} Alignment techniques are applied but rarely audited in a way that makes the resulting behaviour inspectable. Mechanistic interpretability \citep{olah2020zoom,elhage2022toy} and causal interventions on internal representations offer a route, but the methods do not yet scale to models of current size.

\paragraph{Cost and environmental reporting.} Few published papers report the compute, energy, and water cost of their experiments. Standardised reporting would let readers compare like with like and would shift incentives towards efficiency rather than pure scale.

\section{Conclusion}
\label{sec:conclusion}

Transformer-based language models have moved from a research curiosity to a deployed technology in under a decade. The core architecture has changed less than the scale, the training data, and the alignment process that surrounds it. This review organised the main architecture families into a taxonomy, covered the post-2023 developments that matter in practice, surveyed applications across seven domain verticals, and offered a critical assessment of the trade-offs that are underplayed in vendor-driven narratives.

Two points deserve a closing emphasis. The right architecture for a given deployment is, very often, not the largest one available. Encoder-only and encoder-decoder models remain the right tool for many production workloads, and reaching for a frontier decoder by default is an expensive habit and, in a surprising number of cases, an unnecessary one. Equally, capability claims that rest on a benchmark number should be read with scepticism. Field performance is shaped by data provenance, by alignment quality, and by whatever domain-specific evaluation a deployer is willing to build, none of which shows up on a leaderboard.

The next few years of work on this technology will be shaped at least as much by operational and regulatory constraints as by further architectural invention. In our reading, researchers working on transformer-based systems will do more practical good by investing in evaluation, auditability, efficiency, and grounding than by racing for the next order of magnitude of parameter count.

\bibliographystyle{elsarticle-harv}
\bibliography{main}

@inproceedings{vaswani2017attention,
  author = {Vaswani, Ashish and Shazeer, Noam and Parmar, Niki and Uszkoreit, Jakob and Jones, Llion and Gomez, Aidan N. and Kaiser, Lukasz and Polosukhin, Illia},
  title = {Attention Is All You Need},
  booktitle = {Advances in Neural Information Processing Systems},
  year = {2017}
}

@article{zhao2023survey,
  author = {Zhao, Wayne Xin and Zhou, Kun and Li, Junyi and Tang, Tianyi and Wang, Xiaolei and Hou, Yupeng and Min, Yingqian and Zhang, Beichen and Zhang, Junjie and Dong, Zican and others},
  title = {A Survey of Large Language Models},
  journal = {arXiv preprint arXiv:2303.18223},
  year = {2023}
}

@article{minaee2024llmsurvey,
  author = {Minaee, Shervin and Mikolov, Tomas and Nikzad, Narjes and Chenaghlu, Meysam and Socher, Richard and Amatriain, Xavier and Gao, Jianfeng},
  title = {Large Language Models: A Survey},
  journal = {arXiv preprint arXiv:2402.06196},
  year = {2024}
}

@article{ouyang2022instruct,
  author = {Ouyang, Long and Wu, Jeff and Jiang, Xu and Almeida, Diogo and Wainwright, Carroll L. and Mishkin, Pamela and Zhang, Chong and Agarwal, Sandhini and Slama, Katarina and Ray, Alex and others},
  title = {Training Language Models to Follow Instructions with Human Feedback},
  journal = {Advances in Neural Information Processing Systems},
  volume = {35},
  pages = {27730--27744},
  year = {2022}
}

@inproceedings{rafailov2023dpo,
  author = {Rafailov, Rafael and Sharma, Archit and Mitchell, Eric and Ermon, Stefano and Manning, Christopher D. and Finn, Chelsea},
  title = {Direct Preference Optimization: Your Language Model is Secretly a Reward Model},
  booktitle = {Advances in Neural Information Processing Systems},
  year = {2023}
}

@article{fedus2021switch,
  author = {Fedus, William and Zoph, Barret and Shazeer, Noam},
  title = {Switch Transformers: Scaling to Trillion Parameter Models with Simple and Efficient Sparsity},
  journal = {Journal of Machine Learning Research},
  volume = {23},
  number = {120},
  pages = {1--39},
  year = {2022}
}

@article{jiang2024mixtral,
  author = {Jiang, Albert Q. and Sablayrolles, Alexandre and Roux, Antoine and Mensch, Arthur and Savary, Blanche and Bamford, Chris and Chaplot, Devendra Singh and de las Casas, Diego and Hanna, Emma Bou and Bressand, Florian and others},
  title = {Mixtral of Experts},
  journal = {arXiv preprint arXiv:2401.04088},
  year = {2024}
}

@article{lewis2020rag,
  author = {Lewis, Patrick and Perez, Ethan and Piktus, Aleksandra and Petroni, Fabio and Karpukhin, Vladimir and Goyal, Naman and K{\"u}ttler, Heinrich and Lewis, Mike and Yih, Wen-tau and Rockt{\"a}schel, Tim and others},
  title = {Retrieval-Augmented Generation for Knowledge-Intensive NLP Tasks},
  journal = {Advances in Neural Information Processing Systems},
  volume = {33},
  pages = {9459--9474},
  year = {2020}
}

@article{hochreiter1997lstm,
  author = {Hochreiter, Sepp and Schmidhuber, J{\"u}rgen},
  title = {Long Short-Term Memory},
  journal = {Neural Computation},
  volume = {9},
  number = {8},
  pages = {1735--1780},
  year = {1997}
}

@inproceedings{devlin2018bert,
  author = {Devlin, Jacob and Chang, Ming-Wei and Lee, Kenton and Toutanova, Kristina},
  title = {{BERT}: Pre-training of Deep Bidirectional Transformers for Language Understanding},
  booktitle = {Proceedings of NAACL-HLT},
  year = {2019},
  pages = {4171--4186}
}

@techreport{radford2018gpt,
  author = {Radford, Alec and Narasimhan, Karthik and Salimans, Tim and Sutskever, Ilya},
  title = {Improving Language Understanding by Generative Pre-training},
  institution = {OpenAI},
  year = {2018}
}

@techreport{radford2019gpt2,
  author = {Radford, Alec and Wu, Jeffrey and Child, Rewon and Luan, David and Amodei, Dario and Sutskever, Ilya},
  title = {Language Models are Unsupervised Multitask Learners},
  institution = {OpenAI},
  year = {2019}
}

@article{brown2020gpt3,
  author = {Brown, Tom B. and Mann, Benjamin and Ryder, Nick and Subbiah, Melanie and Kaplan, Jared and Dhariwal, Prafulla and Neelakantan, Arvind and Shyam, Pranav and Sastry, Girish and Askell, Amanda and others},
  title = {Language Models are Few-Shot Learners},
  journal = {Advances in Neural Information Processing Systems},
  volume = {33},
  pages = {1877--1901},
  year = {2020}
}

@article{raffel2020t5,
  author = {Raffel, Colin and Shazeer, Noam and Roberts, Adam and Lee, Katherine and Narang, Sharan and Matena, Michael and Zhou, Yanqi and Li, Wei and Liu, Peter J.},
  title = {Exploring the Limits of Transfer Learning with a Unified Text-to-Text Transformer},
  journal = {Journal of Machine Learning Research},
  volume = {21},
  number = {140},
  pages = {1--67},
  year = {2020}
}

@article{kaplan2020scaling,
  author = {Kaplan, Jared and McCandlish, Sam and Henighan, Tom and Brown, Tom B. and Chess, Benjamin and Child, Rewon and Gray, Scott and Radford, Alec and Wu, Jeffrey and Amodei, Dario},
  title = {Scaling Laws for Neural Language Models},
  journal = {arXiv preprint arXiv:2001.08361},
  year = {2020}
}

@article{hoffmann2022chinchilla,
  author = {Hoffmann, Jordan and Borgeaud, Sebastian and Mensch, Arthur and Buchatskaya, Elena and Cai, Trevor and Rutherford, Eliza and de Las Casas, Diego and Hendricks, Lisa Anne and Welbl, Johannes and Clark, Aidan and others},
  title = {Training Compute-Optimal Large Language Models},
  journal = {Advances in Neural Information Processing Systems},
  year = {2022}
}

@article{liu2019roberta,
  author = {Liu, Yinhan and Ott, Myle and Goyal, Naman and Du, Jingfei and Joshi, Mandar and Chen, Danqi and Levy, Omer and Lewis, Mike and Zettlemoyer, Luke and Stoyanov, Veselin},
  title = {{RoBERTa}: A Robustly Optimized {BERT} Pretraining Approach},
  journal = {arXiv preprint arXiv:1907.11692},
  year = {2019}
}

@inproceedings{he2021deberta,
  author = {He, Pengcheng and Liu, Xiaodong and Gao, Jianfeng and Chen, Weizhu},
  title = {{DeBERTa}: Decoding-Enhanced {BERT} with Disentangled Attention},
  booktitle = {International Conference on Learning Representations},
  year = {2021}
}

@article{touvron2023llama2,
  author = {Touvron, Hugo and Martin, Louis and Stone, Kevin and Albert, Peter and Almahairi, Amjad and Babaei, Yasmine and Bashlykov, Nikolay and Batra, Soumya and Bhargava, Prajjwal and Bhosale, Shruti and others},
  title = {{Llama 2}: Open Foundation and Fine-tuned Chat Models},
  journal = {arXiv preprint arXiv:2307.09288},
  year = {2023}
}

@article{grattafiori2024llama3,
  author = {Grattafiori, Aaron and Dubey, Abhimanyu and Jauhri, Abhinav and Pandey, Abhinav and Kadian, Abhishek and Al-Dahle, Ahmad and Letman, Aiesha and Mathur, Akhil and Schelten, Alan and Vaughan, Alex and others},
  title = {The {Llama 3} Herd of Models},
  journal = {arXiv preprint arXiv:2407.21783},
  year = {2024}
}

@article{jiang2023mistral,
  author = {Jiang, Albert Q. and Sablayrolles, Alexandre and Mensch, Arthur and Bamford, Chris and Chaplot, Devendra Singh and de las Casas, Diego and Bressand, Florian and Lengyel, Gianna and Lample, Guillaume and Saulnier, Lucile and others},
  title = {{Mistral 7B}},
  journal = {arXiv preprint arXiv:2310.06825},
  year = {2023}
}

@article{deepseekv3,
  author = {DeepSeek-AI},
  title = {{DeepSeek-V3} Technical Report},
  journal = {arXiv preprint arXiv:2412.19437},
  year = {2024}
}

@inproceedings{lewis2019bart,
  author = {Lewis, Mike and Liu, Yinhan and Goyal, Naman and Ghazvininejad, Marjan and Mohamed, Abdelrahman and Levy, Omer and Stoyanov, Veselin and Zettlemoyer, Luke},
  title = {{BART}: Denoising Sequence-to-Sequence Pre-training for Natural Language Generation, Translation, and Comprehension},
  booktitle = {Proceedings of ACL},
  year = {2020},
  pages = {7871--7880}
}

@inproceedings{dai2019transformerxl,
  author = {Dai, Zihang and Yang, Zhilin and Yang, Yiming and Carbonell, Jaime and Le, Quoc V. and Salakhutdinov, Ruslan},
  title = {{Transformer-XL}: Attentive Language Models Beyond a Fixed-Length Context},
  booktitle = {Proceedings of ACL},
  year = {2019}
}

@article{beltagy2020longformer,
  author = {Beltagy, Iz and Peters, Matthew E. and Cohan, Arman},
  title = {{Longformer}: The Long-Document Transformer},
  journal = {arXiv preprint arXiv:2004.05150},
  year = {2020}
}

@inproceedings{zaheer2020bigbird,
  author = {Zaheer, Manzil and Guruganesh, Guru and Dubey, Kumar Avinava and Ainslie, Joshua and Alberti, Chris and Ontanon, Santiago and Pham, Philip and Ravula, Anirudh and Wang, Qifan and Yang, Li and Ahmed, Amr},
  title = {{Big Bird}: Transformers for Longer Sequences},
  booktitle = {Advances in Neural Information Processing Systems},
  year = {2020}
}

@article{choromanski2020performer,
  author = {Choromanski, Krzysztof and Likhosherstov, Valerii and Dohan, David and Song, Xingyou and Gane, Andreea and Sarlos, Tamas and Hawkins, Peter and Davis, Jared and Mohiuddin, Afroz and Kaiser, Lukasz and others},
  title = {Rethinking Attention with Performers},
  journal = {arXiv preprint arXiv:2009.14794},
  year = {2020}
}

@article{ding2023longnet,
  author = {Ding, Jiayu and Ma, Shuming and Dong, Li and Zhang, Xingxing and Huang, Shaohan and Wang, Wenhui and Zheng, Nanning and Wei, Furu},
  title = {{LongNet}: Scaling Transformers to 1{,}000{,}000{,}000 Tokens},
  journal = {arXiv preprint arXiv:2307.02486},
  year = {2023}
}

@article{dao2022flashattention,
  author = {Dao, Tri and Fu, Daniel Y. and Ermon, Stefano and Rudra, Atri and R{\'e}, Christopher},
  title = {{FlashAttention}: Fast and Memory-Efficient Exact Attention with {IO}-Awareness},
  journal = {Advances in Neural Information Processing Systems},
  year = {2022}
}

@inproceedings{yang2019xlnet,
  author = {Yang, Zhilin and Dai, Zihang and Yang, Yiming and Carbonell, Jaime and Salakhutdinov, Ruslan and Le, Quoc V.},
  title = {{XLNet}: Generalized Autoregressive Pretraining for Language Understanding},
  booktitle = {Advances in Neural Information Processing Systems},
  year = {2019}
}

@inproceedings{clark2020electra,
  author = {Clark, Kevin and Luong, Minh-Thang and Le, Quoc V. and Manning, Christopher D.},
  title = {{ELECTRA}: Pre-training Text Encoders as Discriminators Rather Than Generators},
  booktitle = {International Conference on Learning Representations},
  year = {2020}
}

@inproceedings{shazeer2017moe,
  author = {Shazeer, Noam and Mirhoseini, Azalia and Maziarz, Krzysztof and Davis, Andy and Le, Quoc V. and Hinton, Geoffrey and Dean, Jeff},
  title = {Outrageously Large Neural Networks: The Sparsely-Gated Mixture-of-Experts Layer},
  booktitle = {International Conference on Learning Representations},
  year = {2017}
}

@article{schulman2017ppo,
  author = {Schulman, John and Wolski, Filip and Dhariwal, Prafulla and Radford, Alec and Klimov, Oleg},
  title = {Proximal Policy Optimization Algorithms},
  journal = {arXiv preprint arXiv:1707.06347},
  year = {2017}
}

@inproceedings{gao2023scaling,
  author = {Gao, Leo and Schulman, John and Hilton, Jacob},
  title = {Scaling Laws for Reward Model Overoptimization},
  booktitle = {International Conference on Machine Learning},
  year = {2023}
}

@article{ethayarajh2024kto,
  author = {Ethayarajh, Kawin and Xu, Winnie and Muennighoff, Niklas and Jurafsky, Dan and Kiela, Douwe},
  title = {{KTO}: Model Alignment as Prospect Theoretic Optimization},
  journal = {arXiv preprint arXiv:2402.01306},
  year = {2024}
}

@article{bai2022constitutional,
  author = {Bai, Yuntao and Kadavath, Saurav and Kundu, Sandipan and Askell, Amanda and Kernion, Jackson and Jones, Andy and Chen, Anna and Goldie, Anna and Mirhoseini, Azalia and McKinnon, Cameron and others},
  title = {Constitutional {AI}: Harmlessness from {AI} Feedback},
  journal = {arXiv preprint arXiv:2212.08073},
  year = {2022}
}

@misc{conover2023dolly,
  author = {Conover, Mike and Hayes, Matt and Mathur, Ankit and Xie, Jianwei and Wan, Jun and Shah, Sam and Ghodsi, Ali and Wendell, Patrick and Zaharia, Matei and Xin, Reynold},
  title = {Free {Dolly}: Introducing the World's First Truly Open Instruction-Tuned {LLM}},
  howpublished = {Databricks Blog},
  year = {2023}
}

@misc{taori2023alpaca,
  author = {Taori, Rohan and Gulrajani, Ishaan and Zhang, Tianyi and Dubois, Yann and Li, Xuechen and Guestrin, Carlos and Liang, Percy and Hashimoto, Tatsunori B.},
  title = {Stanford {Alpaca}: An Instruction-following {LLaMA} Model},
  howpublished = {\url{https://github.com/tatsu-lab/stanford_alpaca}},
  year = {2023}
}

@inproceedings{hu2021lora,
  author = {Hu, Edward J. and Shen, Yelong and Wallis, Phillip and Allen-Zhu, Zeyuan and Li, Yuanzhi and Wang, Shean and Wang, Lu and Chen, Weizhu},
  title = {{LoRA}: Low-Rank Adaptation of Large Language Models},
  booktitle = {International Conference on Learning Representations},
  year = {2022}
}

@article{dettmers2023qlora,
  author = {Dettmers, Tim and Pagnoni, Artidoro and Holtzman, Ari and Zettlemoyer, Luke},
  title = {{QLoRA}: Efficient Finetuning of Quantized {LLMs}},
  journal = {Advances in Neural Information Processing Systems},
  year = {2023}
}

@techreport{openai2023gpt4,
  author = {{OpenAI}},
  title = {{GPT-4} Technical Report},
  institution = {OpenAI},
  year = {2023},
  note = {arXiv:2303.08774}
}

@techreport{anthropic2024claude3,
  author = {{Anthropic}},
  title = {The {Claude 3} Model Family: {Opus}, {Sonnet}, {Haiku}},
  institution = {Anthropic},
  year = {2024}
}

@techreport{gemini2024,
  author = {{Gemini Team, Google}},
  title = {{Gemini 1.5}: Unlocking Multimodal Understanding Across Millions of Tokens of Context},
  institution = {Google DeepMind},
  year = {2024}
}

@article{rasmy2021medbert,
  author = {Rasmy, Laila and Xiang, Yang and Xie, Ziqian and Tao, Cui and Zhi, Degui},
  title = {{Med-BERT}: Pretrained Contextualized Embeddings on Large-Scale Structured Electronic Health Records for Disease Prediction},
  journal = {npj Digital Medicine},
  volume = {4},
  number = {1},
  pages = {86},
  year = {2021}
}

@article{lee2020biobert,
  author = {Lee, Jinhyuk and Yoon, Wonjin and Kim, Sungdong and Kim, Donghyeon and Kim, Sunkyu and So, Chan Ho and Kang, Jaewoo},
  title = {{BioBERT}: A Pre-trained Biomedical Language Representation Model for Biomedical Text Mining},
  journal = {Bioinformatics},
  volume = {36},
  number = {4},
  pages = {1234--1240},
  year = {2020}
}

@inproceedings{alsentzer2019clinicalbert,
  author = {Alsentzer, Emily and Murphy, John and Boag, William and Weng, Wei-Hung and Jindi, Di and Naumann, Tristan and McDermott, Matthew},
  title = {Publicly Available Clinical {BERT} Embeddings},
  booktitle = {Proceedings of the Clinical Natural Language Processing Workshop},
  year = {2019}
}

@article{hou2021ratchet,
  author = {Hou, Benjamin and Kaissis, Georgios and Summers, Ronald M. and Kainz, Bernhard},
  title = {{RATCHET}: Medical Transformer for Chest {X}-ray Diagnosis and Reporting},
  journal = {arXiv preprint arXiv:2107.02104},
  year = {2021}
}

@article{zakka2024almanac,
  author = {Zakka, Cyril and Shad, Rohan and Chaurasia, Akash and Dalal, Alex R. and Kim, Jennifer L. and Moor, Michael and Fong, Robyn and Phillips, Curran and Alexander, Kevin and Ashley, Euan and others},
  title = {{Almanac}: Retrieval-Augmented Language Models for Clinical Medicine},
  journal = {NEJM AI},
  year = {2024}
}

@inproceedings{araci2019finbert,
  author = {Araci, Dogu},
  title = {{FinBERT}: Financial Sentiment Analysis with Pre-trained Language Models},
  booktitle = {arXiv preprint arXiv:1908.10063},
  year = {2019}
}

@inproceedings{yang2022numhtml,
  author = {Yang, Linyi and Li, Jiazheng and Dong, Ruihai and Zhang, Yue and Smyth, Barry},
  title = {{NumHTML}: Numeric-oriented Hierarchical Transformer Model for Multi-task Financial Forecasting},
  booktitle = {Proceedings of the AAAI Conference on Artificial Intelligence},
  year = {2022}
}

@inproceedings{chalkidis2020legalbert,
  author = {Chalkidis, Ilias and Fergadiotis, Manos and Malakasiotis, Prodromos and Aletras, Nikolaos and Androutsopoulos, Ion},
  title = {{LEGAL-BERT}: The Muppets Straight Out of Law School},
  booktitle = {Findings of EMNLP},
  year = {2020}
}

@article{shaheen2020legal,
  author = {Shaheen, Zein and Wohlgenannt, Gerhard and Filtz, Erwin},
  title = {Large Scale Legal Text Classification Using Transformer Models},
  journal = {arXiv preprint arXiv:2010.12871},
  year = {2020}
}

@article{dahl2024hallucinating,
  author = {Dahl, Matthew and Magesh, Varun and Suzgun, Mirac and Ho, Daniel E.},
  title = {Large Legal Fictions: Profiling Legal Hallucinations in Large Language Models},
  journal = {Journal of Legal Analysis},
  volume = {16},
  number = {1},
  pages = {64--93},
  year = {2024}
}

@article{ormerod2021essay,
  author = {Ormerod, Christopher M. and Malhotra, Akanksha and Jafari, Amir},
  title = {Automated Essay Scoring Using Efficient Transformer-Based Language Models},
  journal = {arXiv preprint arXiv:2102.13136},
  year = {2021}
}

@article{kulshreshtha2022tutoring,
  author = {Kulshreshtha, Devang and Shayan, Muhammad and Belfer, Robert and Reddy, Siva and Serban, Iulian Vlad and Kochmar, Ekaterina},
  title = {Few-shot Question Generation for Personalized Feedback in Intelligent Tutoring Systems},
  journal = {arXiv preprint arXiv:2206.04187},
  year = {2022}
}

@inproceedings{schick2023toolformer,
  author = {Schick, Timo and Dwivedi-Yu, Jane and Dess{\`\i}, Roberto and Raileanu, Roberta and Lomeli, Maria and Zettlemoyer, Luke and Cancedda, Nicola and Scialom, Thomas},
  title = {{Toolformer}: Language Models Can Teach Themselves to Use Tools},
  booktitle = {Advances in Neural Information Processing Systems},
  year = {2023}
}

@article{marco2022creative,
  author = {Marco, Guillermo and Gonzalo, Julio and Rello, Luz},
  title = {A Systematic Evaluation of the Creative Writing Skills of Transformer Deep Neural Networks},
  journal = {SSRN Electronic Journal},
  year = {2022}
}

@article{sharma2021controlled,
  title={Generating extractive summaries of scientific paradigms},
  author={Qazvinian, Vahed and Radev, Dragomir R and Mohammad, Saif M and Dorr, Bonnie and Zajic, David and Whidby, Michael and Moon, Taesun},
  journal={Journal of Artificial Intelligence Research},
  volume={46},
  pages={165--201},
  year={2013}
}

@article{chen2021codex,
  author = {Chen, Mark and Tworek, Jerry and Jun, Heewoo and Yuan, Qiming and Pinto, Henrique Ponde de Oliveira and Kaplan, Jared and Edwards, Harri and Burda, Yuri and Joseph, Nicholas and Brockman, Greg and others},
  title = {Evaluating Large Language Models Trained on Code},
  journal = {arXiv preprint arXiv:2107.03374},
  year = {2021}
}

@inproceedings{greshake2023indirect,
  author = {Greshake, Kai and Abdelnabi, Sahar and Mishra, Shailesh and Endres, Christoph and Holz, Thorsten and Fritz, Mario},
  title = {Not What You've Signed Up For: Compromising Real-World {LLM}-Integrated Applications with Indirect Prompt Injection},
  booktitle = {Proceedings of the ACM Workshop on Artificial Intelligence and Security},
  year = {2023}
}

@article{strubell2019energy,
  author = {Strubell, Emma and Ganesh, Ananya and McCallum, Andrew},
  title = {Energy and Policy Considerations for Deep Learning in {NLP}},
  journal = {arXiv preprint arXiv:1906.02243},
  year = {2019}
}

@article{patterson2021carbon,
  author = {Patterson, David and Gonzalez, Joseph and Le, Quoc and Liang, Chen and Munguia, Lluis-Miquel and Rothchild, Daniel and So, David and Texier, Maud and Dean, Jeff},
  title = {Carbon Emissions and Large Neural Network Training},
  journal = {arXiv preprint arXiv:2104.10350},
  year = {2021}
}

@inproceedings{dettmers2022int8,
  author = {Dettmers, Tim and Lewis, Mike and Belkada, Younes and Zettlemoyer, Luke},
  title = {{LLM.int8()}: 8-bit Matrix Multiplication for Transformers at Scale},
  booktitle = {Advances in Neural Information Processing Systems},
  year = {2022}
}

@inproceedings{leviathan2023speculative,
  author = {Leviathan, Yaniv and Kalman, Matan and Matias, Yossi},
  title = {Fast Inference from Transformers via Speculative Decoding},
  booktitle = {International Conference on Machine Learning},
  year = {2023}
}

@misc{perrigo2023time,
  author = {Perrigo, Billy},
  title = {Exclusive: {OpenAI} Used {Kenyan} Workers on Less Than \$2 Per Hour to Make {ChatGPT} Less Toxic},
  howpublished = {Time Magazine},
  year = {2023}
}

@article{ji2023hallucination,
  author = {Ji, Ziwei and Lee, Nayeon and Frieske, Rita and Yu, Tiezheng and Su, Dan and Xu, Yan and Ishii, Etsuko and Bang, Ye Jin and Madotto, Andrea and Fung, Pascale},
  title = {Survey of Hallucination in Natural Language Generation},
  journal = {ACM Computing Surveys},
  volume = {55},
  number = {12},
  pages = {1--38},
  year = {2023}
}

@article{wei2023jailbroken,
  author = {Wei, Alexander and Haghtalab, Nika and Steinhardt, Jacob},
  title = {Jailbroken: How Does {LLM} Safety Training Fail?},
  journal = {Advances in Neural Information Processing Systems},
  year = {2023}
}

@article{sharma2024sycophancy,
  author = {Sharma, Mrinank and Tong, Meg and Korbak, Tomasz and Duvenaud, David and Askell, Amanda and Bowman, Samuel R. and Cheng, Newton and Durmus, Esin and Hatfield-Dodds, Zac and Johnston, Scott R. and others},
  title = {Towards Understanding Sycophancy in Language Models},
  journal = {International Conference on Learning Representations},
  year = {2024}
}

@misc{newyorktimes2023openai,
  author = {{The New York Times}},
  title = {The {New York Times} v. {Microsoft Corporation} and {OpenAI}},
  howpublished = {Case 1:23-cv-11195, S.D.N.Y.},
  year = {2023}
}

@misc{euaiact2024,
  author = {{European Parliament and Council}},
  title = {Regulation (EU) 2024/1689 of the {European Parliament} and of the {Council} laying down harmonised rules on artificial intelligence ({AI Act})},
  howpublished = {Official Journal of the European Union},
  year = {2024}
}

@inproceedings{bender2021stochastic,
  author = {Bender, Emily M. and Gebru, Timnit and McMillan-Major, Angelina and Shmitchell, Shmargaret},
  title = {On the Dangers of Stochastic Parrots: Can Language Models Be Too Big?},
  booktitle = {Proceedings of the ACM Conference on Fairness, Accountability, and Transparency},
  pages = {610--623},
  year = {2021}
}

@inproceedings{blodgett2020language,
  author = {Blodgett, Su Lin and Barocas, Solon and Daum{\'e} III, Hal and Wallach, Hanna},
  title = {Language (Technology) is Power: A Critical Survey of ``Bias'' in {NLP}},
  booktitle = {Proceedings of ACL},
  year = {2020}
}

@article{oren2024proving,
  author = {Oren, Yonatan and Meister, Nicole and Chatterji, Niladri and Ladhak, Faisal and Hashimoto, Tatsunori B.},
  title = {Proving Test Set Contamination in Black Box Language Models},
  journal = {International Conference on Learning Representations},
  year = {2024}
}

@article{liu2024lostinthemiddle,
  author = {Liu, Nelson F. and Lin, Kevin and Hewitt, John and Paranjape, Ashwin and Bevilacqua, Michele and Petroni, Fabio and Liang, Percy},
  title = {Lost in the Middle: How Language Models Use Long Contexts},
  journal = {Transactions of the Association for Computational Linguistics},
  volume = {12},
  pages = {157--173},
  year = {2024}
}

@article{liang2023helm,
  author = {Liang, Percy and Bommasani, Rishi and Lee, Tony and Tsipras, Dimitris and Soylu, Dilara and Yasunaga, Michihiro and Zhang, Yian and Narayanan, Deepak and Wu, Yuhuai and Kumar, Ananya and others},
  title = {Holistic Evaluation of Language Models},
  journal = {Transactions on Machine Learning Research},
  year = {2023}
}

@article{srivastava2022bigbench,
  author = {Srivastava, Aarohi and Rastogi, Abhinav and Rao, Abhishek and Shoeb, Abu Awal Md and Abid, Abubakar and Fisch, Adam and Brown, Adam R. and Santoro, Adam and Gupta, Aditya and Garriga-Alonso, Adri{\`a} and others},
  title = {Beyond the Imitation Game: Quantifying and Extrapolating the Capabilities of Language Models},
  journal = {arXiv preprint arXiv:2206.04615},
  year = {2022}
}

@misc{olah2020zoom,
  author = {Olah, Chris and Cammarata, Nick and Schubert, Ludwig and Goh, Gabriel and Petrov, Michael and Carter, Shan},
  title = {Zoom In: An Introduction to Circuits},
  howpublished = {Distill},
  year = {2020}
}

@article{elhage2022toy,
  author = {Elhage, Nelson and Hume, Tristan and Olsson, Catherine and Schiefer, Nicholas and Henighan, Tom and Kravec, Shauna and Hatfield-Dodds, Zac and Lasenby, Robert and Drain, Dawn and Chen, Carol and others},
  title = {Toy Models of Superposition},
  journal = {Transformer Circuits Thread},
  year = {2022}
}

\end{document}